\documentclass[times,twocolumn,final]{elsarticle}

\usepackage{medima}
\usepackage{framed,multirow}

\usepackage{amssymb}
\usepackage{latexsym}

\usepackage{url}
\usepackage{xcolor}

\usepackage[colorlinks=true,citecolor=blue,urlcolor=blue]{hyperref}

\usepackage{tabularx,threeparttable}
\usepackage{bm}

\usepackage{bbding}
\usepackage{makecell}
\usepackage{amsmath}
\usepackage{verbatim}
\usepackage{diagbox}
\usepackage{color,graphicx}
\usepackage{arydshln}
\usepackage{epstopdf}
\usepackage{footnote}
\usepackage{subfigure}
\usepackage{ulem}
\usepackage{pifont}
\usepackage{caption}

\definecolor{newcolor}{rgb}{.8,.349,.1}

\begin{document}

\begin{frontmatter}

\title{MA-SAM: Modality-agnostic SAM Adaptation for 3D Medical Image Segmentation}

\author[1]{Cheng Chen}
\author[2]{Juzheng Miao}
\author[1]{Dufan Wu}
\author[3]{Zhiling Yan}
\author[1]{Sekeun Kim}
\author[1]{Jiang Hu}
\author[4,1]{Aoxiao Zhong}
\author[5,1]{Zhengliang Liu}
\author[3]{\\Lichao Sun}
\author[1]{Xiang Li}
\author[5]{Tianming Liu}
\author[2]{Pheng-Ann Heng}
\author[1]{Quanzheng Li\corref{cor1}}
\cortext[cor1]{Corresponding author: li.quanzheng@mgh.harvard.edu}


\address[1]{Center of Advanced Medical Computing and Analysis, Massachusetts General Hospital and Harvard Medical School, Boston, MA 02114, USA}
\address[2]{Department of Computer Science and Engineering, The Chinese University of Hong Kong, Hong Kong, China}
\address[3]{Department of Computer Science and Engineering, Lehigh University, Bethlehem, PA 18015, USA}
\address[4]{Harvard John A. Paulson School of Engineering and Applied Sciences, Harvard University, Cambridge, MA 02138, USA}
\address[5]{School of Computing, The University of Georgia, Athens, GA 30602, USA}

\begin{abstract}
The Segment Anything Model (SAM), a foundation model for general image segmentation, has demonstrated impressive zero-shot performance across numerous natural image segmentation tasks. 
However, SAM's performance significantly declines when applied to medical images, primarily due to the substantial disparity between natural and medical image domains.
To effectively adapt SAM to medical images, it is important to incorporate critical third-dimensional information, i.e., volumetric or temporal knowledge, during fine-tuning. Simultaneously, we aim to harness SAM's pre-trained weights within its original 2D backbone to the fullest extent.
In this paper, we introduce a modality-agnostic SAM adaptation framework, named as MA-SAM, that is applicable to various volumetric and video medical data. 
Our method roots in the parameter-efficient fine-tuning strategy to update only a small portion of weight increments while preserving the majority of SAM's pre-trained weights.
By injecting a series of 3D adapters into the transformer blocks of the image encoder, our method enables the pre-trained 2D backbone to extract third-dimensional information from input data.
The effectiveness of our method has been comprehensively evaluated on four medical image segmentation tasks, by using 10 public datasets across CT, MRI, and surgical video data.
Remarkably, without using any prompt, our method consistently outperforms various state-of-the-art 3D approaches, surpassing nnU-Net by 0.9\%, 2.6\%, and 9.9\% in Dice for CT multi-organ segmentation, MRI prostate segmentation, and surgical scene segmentation respectively.
Our model also demonstrates strong generalization, and excels in challenging tumor segmentation when prompts are used. Our code is available at: \url{https://github.com/cchen-cc/MA-SAM}.

\end{abstract}

\end{frontmatter}


\section{Introduction}

The rise of foundation models~\citep{bommasani2021opportunities} that are trained on vast and diverse datasets has catalyzed a paradigm shift in intelligent model development. 
Driven by their remarkable generalization and few-shot learning capability, it has become increasingly appealing to adapt a pre-trained large model to a diversity of downstream tasks, as opposed to the traditional approach of crafting and training distinct task-specific models from scratch. 
The Segment Anything Model (SAM)~\citep{SAM} is a recently developed visual foundation model for promptable image segmentation, pre-trained over 1 billion masks on 11 million natural images.
Thanks to its large-scale training data and general model architecture, SAM has demonstrated impressive zero-shot performance on various tasks in the context of natural images. 
Given these merits, a natural question arises: can SAM be directly extended to address the critical medical image segmentation tasks, a domain that has been struggling with limited availability of high-quality images and labels essential for training deep models?
However, due to the significant domain gap between natural images and medical images, the latest works on evaluating SAM on medical images have shown that SAM's zero-shot capability, regardless of whether prompts are employed, falls short for direct deployment on medical images~\citep{huang2023segment,he2023accuracy,wald2023sam}.
In these assessments, SAM obtains inferior performance when compared to state-of-the-art (SOTA) medical image segmentation models, and even encounters complete failure in some challenging tasks. 

Based on these evaluations, it becomes evident that fine-tuning is an essential step for applying SAM to medical images. 
But why are we inclined to adapt SAM for medical image tasks? This can be attributed to three potential advantages associated with SAM.
Firstly, SAM's training dataset consists of an extensive collection of images. Acquiring a similarly large-scale training dataset in the context of medical applications is extremely challenging.
Although SAM's training data only comprises natural images, it is not restricted to any specific medical imaging modality. If SAM fine-tuning proves effective for one type of medical imaging, there is a good chance that the same approach could be applicable to other modalities as well. 
Secondly, after fine-tuning, SAM as a pre-trained large models may possess potential for robust generalization, which is of great importance for effectively deploying intelligent models in critical medical applications.
Thirdly, SAM's prompt design provides a convenient solution for semi-automatic segmentation in tackling difficult tasks, such as tumor segmentation.
In these aspects, SAM provides a general-purpose foundation model with the potential to be adapted across diverse medical imaging modalities, offering good generalization capability for both fully-automatic and semi-automatic segmentation.

Efforts to adapt SAM for medical applications are rapidly growing, with the majority of these approaches relying on SAM's prompt design \citep{cheng2023sam,wu2023medical,deng2023sam,dai2023samaug}. 
However, providing suitable prompts for segmenting each object within medical data is non-trivial. 
For example, consider an abdominal CT volume containing multiple organs, even providing a basic point prompt for each organ in every slice demands substantial efforts. 
Moreover, in cases where segmentation objects present relatively regular shapes and locations, automatic segmentation methods already obtain encouraging results, obviating the need for prompts in semi-automatic segmentation.
In the context of SAM adaptation for automatic medical image segmentation, some recent studies employ parameter-efficient transfer learning (PETL) techniques, such as LoRA~\citep{lora} or Adapters~\citep{houlsby2019parameter}, showing promising performance in automatic segmentation \citep{SAMed,wang2023sam}. 
However, these methods focus on pure 2D adaptation, overlooking the valuable third-dimensional information inherently present in medical images. This includes the crucial 3D spatial information in medical volumetric data and the temporal information in medical video data.

In this paper, we propose a modality-agnostic SAM adaptation method for medical image segmentation, named as MA-SAM, which efficiently and effectively captures the volumetric or temporal information in medical data. 
For the fine-tuning of image encoder, we leverage the PETL technique called FacT \citep{jie2023fact}, which is based on tensorization-decomposition to enhance the tuning efficiency. 
Such fine-tuning approach retains the pre-trained weights to a large extent and only updates lightweight weight increments, ensuring the preservation of general knowledge necessary for object segmentation and reducing the number of parameters that need to be adjusted. 
To bridge the gap between 2D natural images and volumetric or video medical data, we further incorporate a set of 3D adapters into each transformer block of the image encoder to extract the valuable third-dimensional information. 
For the adaptation of the lightweight mask decoder, we employ full fine-tuning and modify its original architecture with a simple yet effective progressive up-sampling mechanism to recover the prediction resolution.
We demonstrate the efficacy of our SAM adaptation framework on multiple medical imaging modalities in tackling various segmentation tasks.
By comparing with multiple SOTA methods, our automatic segmentation demonstrates superior performance and remarkable generalization capability. 
Our main contributions are highlighted as follows:
\begin{itemize}
\item We propose a parameter-efficient fine-tuning method to adapt SAM to volumetric and video medical data. Our method effectively incorporates the essential third-dimensional information from medical images into the 2D network backbone via lightweight 3D adapters.

\item We demonstrate that our SAM adaptation can be applied to various medical imaging modalities, including CT, MRI, and surgical video data, for anatomy, surgical scene, and tumor segmentation. Without using any prompt, our automatic segmentation consistently outperforms competitive SOTA methods by a large margin.

\item We validate that after fine-tuning on medical images, the obtained models present outstanding generalization capability, showing even superior performance than SOTA domain generalization approaches.

\item We show that by further leveraging prompts, our method achieves impressive results in challenging tumor segmentation task, surpassing nnU-Net by 38.7\% in Dice score. 
\end{itemize}

\section{Related work}
\subsection{Vision foundation models}
Foundation models has recently been actively developed in computer vision, although to a lesser extent compared to their prevalence in natural language processing.
Pioneering vision foundation models learn directly from vast image-text pairs sourced from the web in a self-supervised manner. 
Representative works CLIP~\citep{CLIP} and ALIGN~\citep{ALIGN} leverage contrastive learning techniques to train both text and image encoders. 
However, these models primarily excel in tasks that involve mapping images to text, such as classification.
Later on, Florence~\citep{Florence} incorporates universal visual-language representations, showing adaptability to more diverse computer vision tasks. 
One of the latest developments is SAM~\citep{SAM}, a vision foundation model for general-purpose image segmentation. 
By pre-training on 1 billion masks, SAM demonstrates impressive zero-shot capability across numerous image segmentation tasks. 
Concurrently, SegGPT~\citep{SegGPT} and SEEM~\citep{SEEM} have also emerged for general image segmentation, but are pre-trained on relatively smaller datasets compared to SAM.

\subsection{Parameter-efficient transfer learning}
With the remarkable performance exhibited by large models, the paradigm of pre-training large foundation models and subsequently fine-tuning for specific downstream tasks has gained increasing popularity. 
As the pre-trained large models continue to grow in scale, the research on PETL has emerged to achieve effective and efficient adaptation by optimizing only a small subset of model parameters while keeping substantial amount of parameters fixed. 
PETL techniques has been originally proposed in natural language processing, and can be categorized into three main groups~\citep{lialin2023scaling}, including additive methods, selective methods, and reparameterization-based methods. 
Additive methods, such as Adapters \citep{houlsby2019parameter}, aim to augment the existing pre-trained model by introducing additional parameters or layers, and then fine-tuning only these newly introduced components \citep{DBLP:conf/iclr/HeZMBN22,liu2023gpt}.
Selective methods focus on updating a few selected influential layers or internal structure within the model \citep{DBLP:conf/emnlp/Gheini0M21,DBLP:conf/acl/ZakenGR22}.
Reparameterization-based methods, such as LoRA \citep{lora} and FacT \citep{jie2023fact}, leverage low-rank representations to minimize the number of trainable parameters, demonstrating robust and SOTA performance across various PETL tasks. 
Recently, PETL has also been actively studied in computer vision, enabling the effective adaptation of vision foundation models to a wide range of downstream tasks \citep{zhou2022conditional,jia2022visual,pan2022st,wang2023med}.

\subsection{Adapting SAM in medical imaging}

Attracted by SAM's outstanding zero-shot performance in natural images, a plethora of evaluation studies quickly emerged in various medical image segmentation tasks~\citep{huang2023segment,he2023accuracy,wald2023sam,zhou2023can,deng2023segment,hu2023sam,cheng2023sam,zhang2023segment}.
However, due to the large domain gap between natural and medical images, directly applying SAM to medical applications typically resulted in unsatisfactory performance. 
For example, He et al.~\citep{he2023accuracy} assessed SAM's segmentation accuracy on 12 medical imaging datasets and observed that SAM's zero-shot performance lagged significantly behind models trained on domain-specific medical images, with performance gap as large as 70\% in Dice in some tasks. 
Similar observations were reported in \citep{huang2023segment}, even when using different types of prompts. 
These findings suggest the necessity of task-specific fine-tuning to adapt SAM for medical images for a better segmentation performance.

Subsequently, attention has shifted from evaluation to adaptation of SAM to medical images \citep{SAMed,biswas2023polyp,wu2023self,li2023auto,feng2023cheap}.
Driven by the improvements observed with the use of prompts, a majority of works leverage SAM's prompt design during fine-tuning \citep{cheng2023sam,deng2023sam,dai2023samaug,yue2023surgicalsam}.
For instance, SAM-Med2D \citep{cheng2023sam} adopted more comprehensive prompts involving points, bounding boxes, and masks to tailor SAM for 2D medical images, and conducted comprehensive evaluations. 
MSA \citep{wu2023medical} employed point prompts and the Adapter technique to integrate medical domain knowledge into the SAM model.
However, creating prompts for each 2D slice of 3D medical data is labor-intensive. 
In the case of SAM adaptation for fully automatic medical image segmentation \citep{hu2023efficiently,paranjape2023adaptivesam}, SAMed \citep{SAMed} and Wang et al. \citep{wang2023sam} adopted LoRA for fine-tuning, showing superior performance than multiple 2D medical image segmentation methods. 
However, these methods do not take into account the critical 3D volumetric or temporal information, which is well-known to be valuable for enhancing medical image segmentation performance.

\begin{figure*}[!t]
	\centering
	\includegraphics[width=0.99\textwidth]{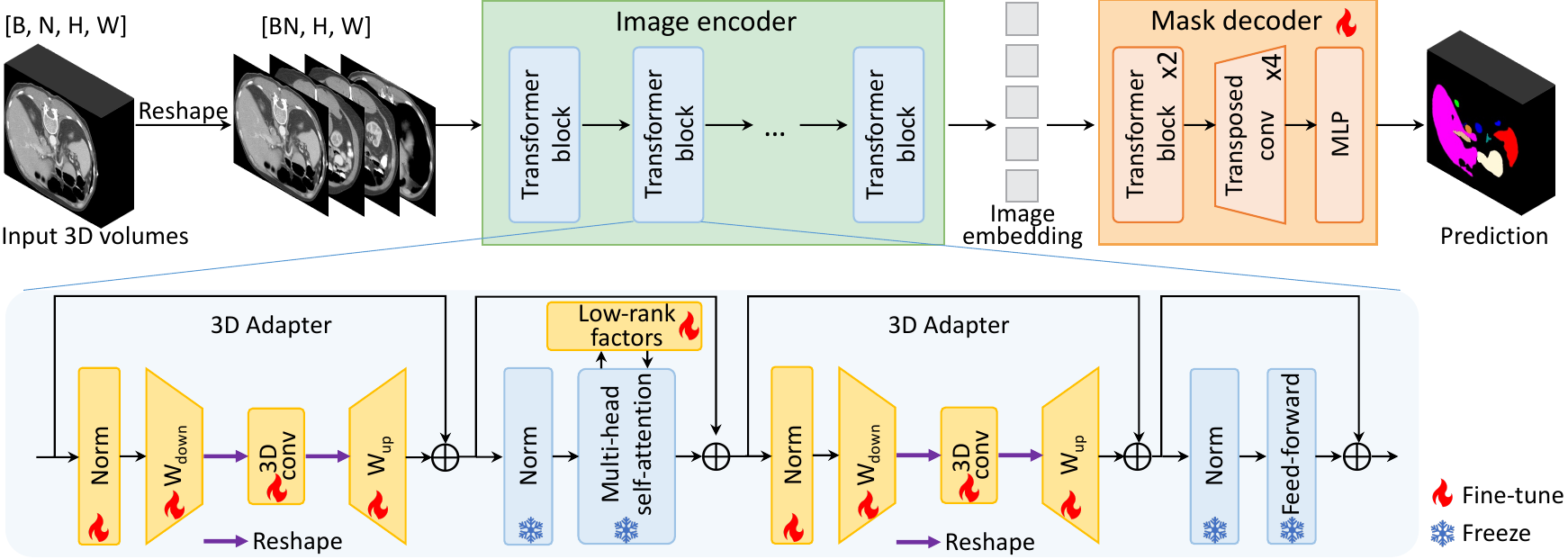}
 \vspace{-0mm}
	\caption{The overview of our proposed modality-agnostic SAM adaptation framework (MA-SAM) for medical image segmentation. The image encoder is updated through a parameter-efficient fine-tuning strategy with FacT. The volumetric or temporal information is effectively incorporated via a set of 3D adapters. The mask decoder is fully fine-tuned and modified to recover the prediction resolution. Reshape operations are used to make 3D operations compatible with the 2D backbone.}
	\label{fig:method}
\end{figure*}
\section{Methodology}
In this section, we first briefly introduce the overview of SAM architecture, then introduce our method for the parameter-efficient fine-tuning of image encoder, the incorporation of volumetric or temporal information, and the adaptation of mask decoder, respectively. An overview of our framework for effective SAM adaptation is illustrated in Fig.~\ref{fig:method}.

\subsection{Overview of SAM}
SAM is a promptable segmentation architecture consisting of three main components, i.e., the image encoder, the prompt encoder, and the mask decoder.
The image encoder employs the Vision Transformer (ViT)~\citep{dosovitskiy2020image} as the backbone, extracting essential features of the images with  a set of transformer blocks.
The prompt encoder takes in various types of prompts including points, boxes, or texts, and encodes these inputs into prompts embeddings to facilitate the segmentation task. 
The mask decoder is designed to be lightweight, which computes the cross-attention between embeddings of image and prompts, and utilizes transposed convolutional layers and multi-layer perception to generate segmentation masks. 
When applying to medical images, the model's performance largely degrades since medical images present distinct texture and objects from natural images.
This highlights the necessity for task-specific fine-tuning of SAM to address such challenges.

\subsection{Parameter-efficient fine-tuning of image encoder}
In order to effectively extract image features, SAM's image encoder comprises a substantial portion of network parameters.
Fine-tuning all these weights is computationally intensive. 
Previous research has shown that PETL techniques can achieve adaptation performance similar to full fine-tuning but with significantly fewer network parameters updated \citep{lora,pan2022st}.
In this regard, we adopt FacT \citep{jie2023fact}, a SOTA PETL technique, that can obtain comparable or superior performance compared to other PETL methods while introducing a smaller number of trainable parameter. 

Based on the common observations that the transformer-based models tend to be redundant in rank, FacT assumes the dense weight increment matrices $\Delta \bm{W}$ used for fine-tuning can be approximated by a set of low-rank factors with cross-layer weight sharing. 
Following the tensor decomposition in FacT, we decompose the weight increments $\Delta \bm{W}$ for each layer into three factors $\bm{U}\in \mathbb{R}^{d\times r}$, $\bm{V}\in \mathbb{R}^{d\times r}$, and $\bm{\Sigma} \in \mathbb{R}^{r\times r}$, where $d$ denotes the feature dimensions in ViT, $r$ stands for the rank of these factors with $r<<d$. It is worth noting that the two factors, $\bm{U}$ and $\bm{V}$, are shared across all layers, while the factor $\bm{\Sigma}$ is unique for each layer. The weight increments can then be calculated using the following equation: 
\begin{equation}
    \Delta \bm{W}_{j,k}=s \cdot \sum_{t_1=1}^{r} \sum_{t_2=1}^{r} \bm{\Sigma}_{t_1,t_2} \bm{U}_{j,t_1} \bm{V}_{k,t_2},
\end{equation}
where $s$ denotes a hyper-parameter for adjusting the learning rate of factors. We fix $s$ as 1 in our experiments and tune the overall learning rate with the optimizer to achieve a similar scaling effect. 
The FacT weight increments are applied to the query and value transformations within each transformer block, while all the other weights initialized from SAM remain frozen, as empirically there were no obvious improvements observed when applying FacT  to other layers. With the FacT weight increments, the query and value transformations become:
\begin{equation}
    \bm{W}_{q/v} = \bm{W}_0 + s \cdot \bm{U} \bm{\Sigma}_{q/v} \bm{V}^{T},
\end{equation}
where $\bm{W}_{q/v}$ denotes the query or value transformation after fine-tuning, $\bm{W}_0$ represents the SAM pre-trained weights.

\subsection{Incorporating volumetric or temporal information}
SAM is initially pre-trained on 2D images, yet medical imaging typically involves more than two dimensions. For example, volumetric CT and MRI data contain crucial 3D spatial information for depicting anatomical structures or lesions, and surgical video data possesses valuable temporal relations between frames. Incorporating this volumetric or temporal knowledge inherent in medical imaging data, is pivotal for the successful transfer learning of SAM in medical applications.
To address this key challenge, we propose to integrate a series of 3D adapters into the 2D transformer blocks within the SAM architecture. These adapters serve the purpose of extracting the essential volumetric or temporal insights needed for medical image analysis. By incorporating these adapters, we bridge the gap between the inherent complexities of medical imaging data and SAM's pre-trained 2D backbone, enabling it to effectively handle multidimensional medical data.

Specifically, as shown in Fig.~\ref{fig:method}, each 3D adapter consists of a normalization layer, a linear down-projection layer, a 3D convolutional layer followed by an activation layer, and a linear up-projection layer. 
The core extraction of volumetric or temporal information primarily resides within the 3D convolutional layer. 
The purpose of the down-projection layer is to reduce the dimensionality of the original $d$-dimensional features into a more compact $c$-dimensional representation, so as to control the number of newly introduced parameters. Conversely, the up-projection layer restores the feature dimensions. 
With $\bm{M}$ denoting feature maps, the 3D adapter can be expressed as:
\begin{equation}
\mathrm{3DAdapter}(\bm{M}) =\bm{M} + \sigma(\mathrm{Conv3D}(\mathrm{Norm}(\bm{M}) \cdot \bm{W}_{\mathrm{down}})) \bm{W}_{\mathrm{up}},
\end{equation}
where Norm denotes the layer normalization, $\sigma$ denotes the activation function, $\bm{W}_\mathrm{down} \in \mathbb{R}^{d\times c}$ and $\bm{W}_\mathrm{up} \in \mathbb{R}^{c\times d}$ denote the linear down- and up-projection layer respectively, and Conv3D denotes the 3D convolutional layer with a kernel size of $3\times 1\times 1$ to specifically extract the third dimensional information. 

To make the 3D adapters compatible with the 2D SAM backbone, for the network inputs, we extract a set of adjacent slices $\bm{x}=\{x_{i-\frac{N-1}{2}},...,x_i,...,x_{i+\frac{N-1}{2}}\}_{i=1}^{B}, \bm{x}\in \mathbb{R}^{B\times N\times H \times W}$. Here, $B$ denotes the batch size, $N$ denotes the number of adjacent slices, and $H\times W$ denotes the slice dimensions.
Before the inputs are passed into the SAM backbone, a reshape operation is applied to transform $\bm{x}\in \mathbb{R}^{B\times N\times H \times W}$ into $\bm{x}\in \mathbb{R}^{BN\times H \times W}$ by merging the adjacent slices into the batch dimension.
Then for the feature maps, prior to feeding into the 3D convolutional layer of a 3D adapter,  they are reshaped from $[BN, H/16, W/16, c]$ to $[B, c, N, H/16, W/16]$. Here $H/16$ and $W/16$ denote the spatial dimensions of feature maps, which are down-sampled by 16 times because of the patch embedding process in transformer. 
After the 3D convolutional operation, the shape of feature maps are changed back again.
In this way, the volumetric or temporal information can be effectively extracted within a 2D network backbone.
For each transformer block, we incorporate two 3D adapters before and after the attention layers, as empirically superior performance can be obtained with such a design.

\subsection{Adapting mask decoder}
The mask decoder within original SAM comprises only two transformer layers, two transposed convolutional layers, and a single multilayer perception layer. 
Considering its lightweight architecture, it is feasible to apply full fine-tuning on the complete mask decoder for effective adaptation on medical images.
During the patch embedding process of the transformer backbone within SAM's image encoder, each $16\times 16$ patch is embedded as a feature vector, leading to $16\times 16$ times down-sampling of the inputs.  
The SAM mask decoder utilizes two consecutive transposed convolutional layers to up-sample the feature maps by 4 times, yet the final predictions generated by SAM remain 4 times lower in resolution than the original shapes. 
Nevertheless, since many anatomical structures or lesions in medical images are quite small, achieving a higher resolution is often necessary to ensure improved discrimination in the context of medical imaging \citep{ronneberger2015u}.

To address this issue, we explore two approaches to tailor the mask decoder for enhanced suitability in medical image segmentation.
For the first approach, termed as ``progressive upsampling", we introduce modest adjustments to the SAM decoder by integrating two additional transposed convolutional operations.
With each layer up-samples the feature maps by a factor of 2, the four transposed convolutional layers progressively restore feature maps to their original input resolution.
The second approach, termed as ``multi-scale fusion", entails creating a design resembling a ``U-shaped" network \citep{ronneberger2015u}. This involves connecting the multi-scale feature maps of the image encoder with corresponding stages of the mask decoder using skip connections, a concept akin to that of U-Net. To achieve this, we uniformly divide the image encoder into four stages, establishing connections between the feature maps of each stage and those of the decoder through a series of up-sampling and convolutional operations.
In our experiments, we have observed that the gradual up-sampling mechanism yields superior outcomes compared to multi-layer feature aggregation, showing the efficacy and simplicity of the progressive up-sampling approach.

\section{Experiments}
We extensively evaluate our method on four medical image segmentation tasks, covering three types of medical imaging modalities from 10 datasets, i.e., abdominal multi-organ segmentation in CT, prostate segmentation in MRI, and surgical scene segmentation in surgical video.
We first conduct comparison with SOTA medical image segmentation methods and SAM fine-tuning methods, and then provide generalization evaluation and in-depth ablation studies to analyze our method.

\subsection{Datasets and evaluation metrics}
\textbf{Task1:} The Beyond the Cranial Vault (BTCV) challenge dataset \citep{landman2015miccai} contains 30 CT volumes with manual annotations for 13 abdominal organs.
Each CT scan contains 85 to 198 slices with the slice thickness varying from 2.5 $mm$ to 5.0 $mm$. 
The axial size is $512\times 512$ for all scans, but with in-plane resolution ranging from 0.54 $\times$ 0.54 $mm^2$ to 0.98 $\times$ 0.98 $mm^2$.
We use the same data split as \citep{swinunetr}, which contains 24 cases for training and 6 cases for testing. 

\textbf{Task2:} We perform prostate segmentation on 6 MRI data sources \citep{liu2020ms}, i.e., Site A to F, that were collected from NIC-ISBI13~\citep{bloch2015nci}, I2CVB~\citep{lemaitre2015computer}, and PROMISE12~\citep{litjens2014evaluation} datasets. 
The case number for each site is 30, 30, 19, 13, 12, 12 respectively, which were randomly divided into 80\% and 20\% for training and testing.
These MRI scans from different sites were acquired with varying imaging protocols and present heterogeneous data distributions, thus were commonly used in previous domain generalization studies \citep{liu2022single}.

\textbf{Task3:} The 2018 MICCAI Robotic Scene Segmentation Challenge (EndoVis18) dataset \citep{allan20202018} comprises 19 sequences, captured using the da Vinci X or Xi system. 
Each sequence contains either 149, 249, or 250 frames at a resolution of $1280\times 1024$. 
The dataset encompasses the surgical scene, with 12 classes annotated for various anatomical structures and robotic instruments.
The dataset is officially split into 15 sequences for training and 4 sequences for testing. 

\renewcommand{\arraystretch}{1.05}
\begin{table*}[!t]
    \centering
    \caption{{Comparison of abdominal multi-organ segmentation results generated from our MA-SAM method and other state-of-the-art methods on BTCV dataset.}}
     \vspace{-2mm}
	    \resizebox{1.0\textwidth}{!}{%
	    \setlength\tabcolsep{1.5pt}
	    \scalebox{0.93}{
	    \begin{tabular}{l|cccccccccccc|c}
	       \hline
	      
	             Methods
	             &Spleen &~~R.Kd~~ &~~L.Kd~ &~~GB~~ &~~Eso.~~ &~Liver~ &Stomach &~Aorta~ &~~IVC~~ &~Veins &Pancreas &AG &Average\\
	      \hline   
	      \hline
       \multicolumn{14}{c}{Dice [\%] $\uparrow$}\\
       \hline
       \hline
	  nnU-Net \citep{isensee2021nnu} &\textbf{97.0} &\textbf{95.3} &{95.3} &63.5 &77.5 &\textbf{97.4} &89.1 &90.1 &\textbf{88.5} &{79.0} &\textbf{87.1} &\textbf{75.2} &{86.3} \\
        3D UX-Net \citep{DBLP:conf/iclr/LeeBHL23} &94.6 &94.2 &94.3 &59.3 &72.2 &96.4 &73.4 &87.2 &84.9 &72.2 &80.9 &67.1 &81.4 \\
        SwinUNETR \citep{tang2022self} &95.6 &94.2 &94.3 &63.6 &75.5 &96.6 &79.2 &89.9 &83.7 &75.0 &82.2 &67.3 &83.1 \\
        nnFormer \citep{zhou2023nnformer} &93.5 &94.9 &95.0 &64.1 &79.5 &96.8 &90.1 &89.7 &85.9 &77.8 &85.6 &73.9 &85.6\\
        
        SAMed\_h \citep{SAMed} &95.3 &92.1 &92.9 &62.1 &75.3 &96.4 &90.2 &87.6 &79.8 &74.2 &77.9 &61.0 &82.1 \\
        \hline
        MA-SAM (Ours) &96.7 &95.1 &\textbf{95.4} &\textbf{68.2} &\textbf{82.1} &96.9 &\textbf{92.8} &\textbf{91.1} &87.5 &\textbf{79.8} &86.6 &73.9 &\textbf{87.2}\\

	  \hline
        \hline
        \multicolumn{14}{c}{HD [\%] $\downarrow$}\\
        \hline
        \hline
        nnU-Net \citep{isensee2021nnu}&1.07 &\textbf{1.19} &1.19 &7.49 &8.56 &\textbf{1.14} &4.84 &14.11 &\textbf{2.87} &5.67 &\textbf{2.31} &\textbf{2.23} &4.39\\
        3D UX-Net \citep{DBLP:conf/iclr/LeeBHL23} &3.17 &1.59 &1.26 &4.53 &13.92 &1.75 &19.72 &12.53 &3.47 &9.99 &3.70 &4.11 &6.68\\
        SwinUNETR \citep{tang2022self} &1.21 &1.41 &1.37 &2.25 &5.82 &1.70 &13.75 &5.92 &4.46 &7.58 &3.53 &3.40 &4.37 \\
        nnFormer \citep{zhou2023nnformer} &78.03 &1.41 &1.43 &3.00 &4.92 &1.38 &4.24 &7.53 &4.02 &6.53 &2.96 &2.76 &9.95 \\
        
        SAMed\_h \citep{SAMed}&1.37 &33.53 &1.84 &6.27 &4.84 &1.77 &7.49 &\textbf{4.97} &7.28 &6.87 &10.00 &6.49 &7.73 \\
        \hline
        MA-SAM (Ours) &\textbf{1.00} &\textbf{1.19} &\textbf{1.07} &\textbf{1.59} &\textbf{3.77} &1.36 &\textbf{3.87} &5.29 &3.12 &\textbf{3.25} &3.93 &2.57 &\textbf{2.67}\\

       \hline
        
\end{tabular}}}
\begin{tablenotes}
\item[*] \footnotesize{SAMed\_h: ViT\_H version of SAMed, R.Kd: Right kidney, L.Kd: Left kidney, GB: Gall ladder, Eso.: Esophagus, IVC: Inferior vena cava, AG: Adrenal gland}
\end{tablenotes}
\label{tab:comparison_btcv}
\end{table*}

\renewcommand{\arraystretch}{1.05}
\begin{table*}[!h]
\centering
\caption{{Comparison of prostate segmentation results generated from our MA-SAM method and other state-of-the-art methods on six prostate MRI datasets.}}
\vspace{-2mm}
\resizebox{1.0\textwidth}{!}{%
\setlength\tabcolsep{2.0pt}
\scalebox{0.93}{
    \begin{tabular}{l|cccccc|c||cccccc|c}
    \hline
          
        Methods &Site A &Site B &Site C &Site D &Site E &Site F &Average &Site A &Site B &Site C &Site D &Site E &Site F &Average\\
        \hline   
        \hline
        &\multicolumn{7}{c||}{Dice [\%] $\uparrow$} &\multicolumn{7}{c}{HD [\%] $\downarrow$}\\
        \hline
        \hline
        
        nnU-Net \citep{isensee2021nnu}   &93.3 &89.2 &89.5 &86.5 &91.0 &90.2 &90.0 &1.74 &2.34 &3.61 &2.98 &2.74 &1.80 &2.54\\
        3D UX-Net \citep{DBLP:conf/iclr/LeeBHL23} &91.8 &86.0 &88.3 &70.4 &85.9 &88.4 &85.1 &1.95 &3.20 &4.37 &9.61 &5.07 &2.67 &4.48 \\
        SwinUNETR \citep{tang2022self}   &88.7 &88.0 &88.4 &71.5 &84.7 &84.6 &84.3 &3.27 &3.02 &4.37 &8.59 &5.24 &2.82 &4.55\\
        nnFormer \citep{zhou2023nnformer} &93.6 &90.1 &89.5 &86.8 &91.9 &90.6 &90.4 &1.73 &2.11 &3.54 &2.93 &2.75 &2.08 &2.52\\
        
        SAMed\_h \citep{SAMed} &94.6 &89.5 &88.6 &87.9 &\textbf{92.7} &91.3 &90.8 &1.14 &3.90 &\textbf{3.10} &3.00 &2.61 &1.67 &2.57\\
        \hline
        MA-SAM (Ours) &\textbf{95.3} &\textbf{92.7} &\textbf{90.4} &\textbf{91.3} &\textbf{92.7} &\textbf{93.1} &\textbf{92.6} &\textbf{1.00} &\textbf{1.54} &3.29 &\textbf{1.80} &\textbf{2.56} &\textbf{1.47} &\textbf{1.94}\\
        
        \hline
    \end{tabular}}}
    \begin{tablenotes}
\item[*] \footnotesize{SAMed\_h: ViT\_H version of SAMed}
\end{tablenotes}
    \label{tab:comparison_prostate}
\end{table*}

\begin{figure*}[!t]
	\centering
	\includegraphics[width=0.99\textwidth]{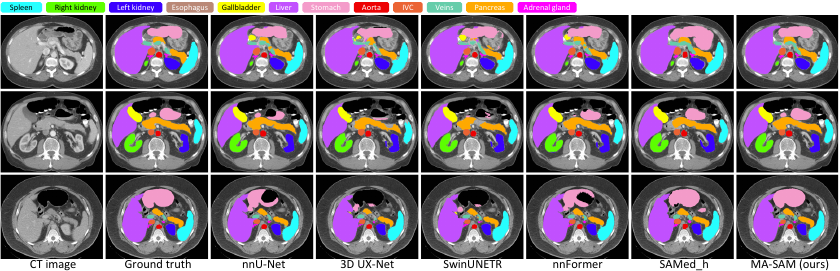}
 \vspace{-2mm}
	\caption{Qualitative visualization of segmentation results generated from our MA-SAM method and other state-of-the-art methods on BTCV dataset. Abdominal organs are denoted in different colors as shown in the corresponding color bar.}
	\label{fig:compare_CT}
\end{figure*}
\textbf{Task4:} The Pancreas Tumor Segmentation task within 2018 MICCAI Medical Segmentation Decathlon Challenge (MSD-Pancreas) dataset \citep{antonelli2022medical} contains 281 CT scans with annotations for pancreas and tumor. 
Each scan comprises 37 to 751 slices with an axial size of 512$\times$512.
We follow \citep{gong20233dsam} to utilize only tumor labels in our experiments and employ the same data split as in their work.

In addition, we use the Multi-Modality Abdominal Multi-Organ Segmentation Challenge (AMOS 22) dataset \citep{ji2022amos} for the evaluation of model generalization. This dataset contains abdominal CT and MRI data that were acquired from different patients. Each scan was annotated with 15 organs, but we focus on the 12 organs that overlap with the BTCV dataset. 300 CT scans and 60 MRI scans in the training and validation sets of AMOS 22 are used for our generalization evaluation.

For data pre-processing, the intensity values of each CT scan in BTCV and MSD-Pancreas datasets were truncated within the range of [-200, 250] Hounsfield Units (HU) and [-50, 200] HU respectively. The intensity of each MRI scan was truncated at the 99th percentile. Each CT or MRI scan was normalized to zero mean and unit variance. 
For surgical video data, each frame was normalized to [0, 1] range. 
We resized all images to $512 \times 512$ for the axial plane of CT and MRI data, as well as for each frame of surgical video sequences. 
For model evaluation, we employ the common metrics, i.e., the Dice score and Hausdorff Distance (HD) to assess pixel-wise segmentation accuracy and the segmentation boundary quality respectively. 
We also report the mean intersection-over-union (mIoU) for the EndoVis18 dataset and the normalized surface distance (NSD) for the MSD-Pancreas dataset to align with previous studies. 

\begin{figure*}[!t]
\begin{minipage}[b]{.5\linewidth}
\centering
\captionsetup{width=.95\linewidth}
\captionof{table}{Comparison of segmentation results from different methods for surgical scene segmentation on Endovis18 dataset.}
\renewcommand{\arraystretch}{1.1}
\resizebox{0.95\textwidth}{!}{%
\setlength\tabcolsep{2.0pt}
    \begin{tabular}{l|cccccc}
    \hline
          
        \multirow{2}{*}{Methods} &\multirow{2}{*}{mIoU} &\multicolumn{4}{c}{Sequence (mIoU)} &\multirow{2}{*}{Dice}\\
        \cline{3-6}
        & &Seq 1 &Seq 2 &Seq 3 &Seq 4 & \\
        \hline   
        \hline

        NCT \citep{shvets2018automatic}       &58.5 &65.8 &55.5 &76.5 &36.2 &- \\
        UNC \citep{ren2020task}       &60.7 &63.3 &57.8 &81.4 &37.3 &- \\
        OTH \citep{chen2018encoder}       &62.1 &69.1 &57.5 &82.9 &39.0 &- \\
        Noisy-LSTM \citep{wang2021noisy} &60.4 &67.0 &56.3 &81.8 &36.4 &69.1 \\
        STswinCL \citep{jin2022exploring}   &63.6 &67.0 &{63.4} &83.7 &40.3 &72.0 \\
        nnU-Net \citep{isensee2021nnu}    &58.7 &65.7 &57.5 &81.3 &30.4 &67.1\\
        SAMed\_h \citep{SAMed} &66.5 &68.7 &60.7 &84.3 &52.3 &74.7 \\
        \hline
        MA-SAM (Ours) &\textbf{69.2} &\textbf{73.4} &\textbf{64.5} &\textbf{85.4} &\textbf{53.4} &\textbf{77.0} \\
        
        \hline
    \end{tabular}}
    \label{tab:comparison_surgical}
\end{minipage}
\begin{minipage}[b]{.495\linewidth}
\centering
\captionsetup{width=.96\linewidth}
\captionof{table}{Comparison of segmentation results from different methods for pancreas tumor segmentation in CT images.}
\renewcommand{\arraystretch}{1.0}
\resizebox{0.95\textwidth}{!}{%
\setlength\tabcolsep{2.0pt}
    \begin{tabular}{l|cc}
    \hline
          
        Methods &Dice $\uparrow$ &NSD $\uparrow$\\
        \hline   
        \hline
        
        nnU-Net \citep{isensee2021nnu}  &41.6   &62.5 \\
        3D UX-Net \citep{DBLP:conf/iclr/LeeBHL23} &34.8   &52.6 \\
        SwinUNETR \citep{tang2022self}  &40.6   &60.0 \\
        nnFormer \citep{zhou2023nnformer}  &36.5   &54.0  \\
        3DSAM-adapter (automatic) \citep{gong20233dsam} &30.2 &45.4 \\
        3DSAM-adapter (10 pts/scan) \citep{gong20233dsam} &57.5 &79.6 \\
        \hline
        MA-SAM (automatic)          &40.2               &59.1 \\
        MA-SAM (1 tight 3D bbx/scan) &\textbf{80.3} &\textbf{97.9} \\
        MA-SAM (1 relaxed 3D bbx/scan) &74.7 &97.1 \\

        \hline
    \end{tabular}}
    \label{tab:comparison_tumor}
\end{minipage}
\end{figure*}

\begin{figure*}[!t]
	\centering
	\includegraphics[width=0.95\textwidth]{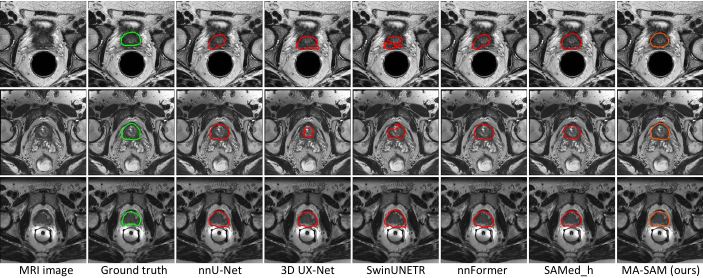}
	\caption{Qualitative visualization of segmentation results generated from our MA-SAM method and other state-of-the-art methods on prostate MRI datasets. The prostate boundary is delineated in green for ground truth, in orange for our method, and in red for other methods, respectively.}
	\label{fig:compare_prostate}
\end{figure*}
\subsection{Implementation details}
The fine-tuning process was supervised using a hybrid segmentation loss, which combines the cross-entropy loss and Dice loss as: $\mathcal{L}_{\text{seg}} = \alpha \mathcal{L}_{\text{ce}} + \beta \mathcal{L}_{\text{Dice}}$. 
The weighting factors $\alpha$ and $\beta$ were set as 0.2 and 0.8 following \citep{SAMed}, except for the surgical video data for which only the Dice loss was utilized. 
Every five consecutive slices were taken as the network inputs.
For data augmentation, we applied a range of transformations including random rotation, flip, erasing, shearing, scaling, translation, posterization, contrast adjustment, brightness modification, and sharpness enhancement.
Our model was trained using Adam optimizer with a batch size of 24. As in \citep{SAMed}, we adopted a warmup training strategy to increase the learning rate linearly to the specific value and then exponentially decrease it towards the end of training to stabilize the training. 
We employed ViT\_H as the backbone of the image encoder and conducted a total of 400 epochs of training, ensuring that the model converged effectively. 
Our framework was implemented in PyTorch 2.0 using 8 NVIDIA A100 GPUs. 

\subsection{Comparison with SOTA methods}
For CT and MRI datasets, we extensively compare our method with various SOTA 3D medical image segmentation methods including CNN-based approaches \textbf{nnU-Net} \citep{isensee2021nnu}, which is a U-Net \citep{ronneberger2015u} based self-configuring framework, showing robust performance on various medical image segmentation competitions, and \textbf{3D UX-Net} \citep{DBLP:conf/iclr/LeeBHL23}, which is a very recent large kernel volumetric ConvNet for 3D medical image segmentation, as well as transformer-based methods \textbf{SwinUNETR} \citep{tang2022self}, which is a 3D transformer-based model with a hierarchical encoder, and \textbf{nnFormer} \citep{zhou2023nnformer}, which is a model combining local and global volume-based self-attention mechanism. 
We also compare our method with the most recent SAM adaptation methods \textbf{SAMed\_h} \citep{SAMed}, which is an automatic 2D medical image segmentation model for organ segmentation, and \textbf{3DSAM-adapter} \citep{gong20233dsam}, which is a promptable 3D medical image segmentation model for tumor segmentation.
For surgical video data, we compare our method with SOTA surgical scene segmentation methods, \textbf{NCT} \citep{shvets2018automatic}, \textbf{UNC} \citep{ren2020task}, and \textbf{OTH} \citep{chen2018encoder}, which are the top-three approaches reported in the challenge, \textbf{Noisy-LSTM} \citep{wang2021noisy} which uses ConvLSTM to learn temporal cues, \textbf{STswinCL} \citep{jin2022exploring} which is a transformer-based model capturing intra- and inter-video relations, and \textbf{nnU-Net}.
For all comparison experiments, the dataset splits remain consistent across all the methods.

\begin{figure}[!t]
\centering
\includegraphics[width=0.455\textwidth]{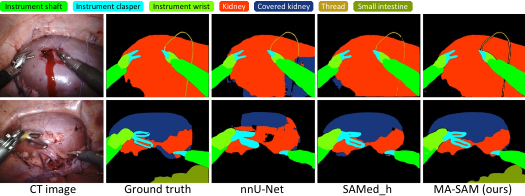}
\vspace{-2mm}
\caption{Qualitative visualization of segmentation results generated from different methods for surgical video data. Classes are denoted in different colors.}
\label{fig:compare_surgical}
\end{figure}

Table~\ref{tab:comparison_btcv} to Table~\ref{tab:comparison_tumor} present comparative results for the four different tasks: abdominal multi-organ segmentation in CT data, prostate MRI segmentation across 6 sites, scene segmentation in surgical video, and tumor segmentation in CT data, respectively.
When prompts are not specified, all methods generate results automatically without using any prompt. 
With our dedicatedly designed fine-tuning strategy for SAM, our method consistently and significantly outperforms other comparison approaches across all the four tasks.
In terms of fully automatic segmentation for the first three tasks, our method improves the Dice score by 0.9\%, 2.6\%, 5\% compared to the second-best performing approach, respectively.
Notably, nnU-Net proves to be a strong competitor, showing robust segmentation performance across CT and MRI datasets. However, in surgical scene segmentation, nnU-Net obtains lower results compared to methods specifically tailored for processing surgical videos.
Our method demonstrates strong performance across both volumetric and video medical data, indicating the potential of unifying the network architecture in these two domains of medical imaging, where previous methods were developed separately.
When comparing with the pure 2D SAM fine-tuning method SAMed\_h, which employs the same network backbone as ours, our method also achieves significantly better results, demonstrating the benefits of incorporating volumetric or temporal information for 3D medical image segmentation. 
The visual comparison results are presented in Fig.~\ref{fig:compare_CT}
to Fig.~\ref{fig:compare_surgical}.

\begin{figure}[!t]
\centering
\includegraphics[width=0.4\textwidth]{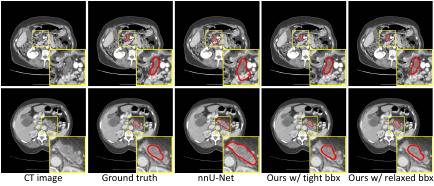}
\vspace{-2mm}
\caption{Qualitative visualization of segmentation results generated from different methods for pancreas tumor segmentation.}
\label{fig:compare_tumor}
\end{figure}

Pancreas tumor segmentation presents a substantial challenge due to the irregular contours and unclear margins of pancreas tumors in CT scans. 
As can be seen in Table~\ref{tab:comparison_tumor} and Fig.~\ref{fig:compare_tumor}, all automatic segmentation models struggle to correctly delineate pancreas tumor regions, obtaining merely a 41.6\% Dice score for the best-performing model. 
We consider in such a demanding segmentation task, the use of prompts become valuable. 
By adding prompts in the form of one tight 3D bounding box per volume into the model, our method remarkably boosts the Dice score from 41.6\% to 80.35\%, demonstrating the effectiveness of leveraging prompts for tumor segmentation.
However, if allowing 5\% relaxation on the tightness of provided bounding box, the performance drops to 74.7\%, showing the effect of prompts quality on segmentation performance.
Our method also significantly outperforms the recent holistic 3D SAM adaptation method 3DSAM-adapter, with 10\% Dice improvement when using automatic segmentation.
This can be attributed to our method's effective incorporation of third-dimensional information into model fine-tuning, as well as its substantial utilization of pre-trained weights from SAM to retain general discriminative knowledge.
We notice that our method's automatic segmentation performance falls slightly behind nnU-Net on tumor segmentation. 
This observation might indicate that SAM fine-tuning might be less effective for objects with ill-defined margins and small sizes, as these characteristics differ from the natural images on which SAM was originally trained.

\subsection{Generalization evaluation}
\begin{figure}[!t]
\centering
\vspace{0mm}
\includegraphics[width=0.49\textwidth]{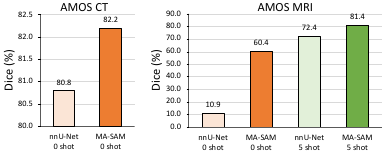}
\vspace{-4mm}
\caption{Comparison of zero-shot and five-shot generalization performance of nnU-Net and our MA-SAM model on AMOS CT and MRI data.}
\label{fig:generalization_CT}
\end{figure}

One of the most appealing advantage of foundation models lies in their impressive generalization capability. 
To investigate the generalization of our models adapted from SAM, we first compare the zero-shot and few-shot capability of nnU-Net and our method by applying models trained on the BTCV CT scans to the AMOS22 CT and MRI scans. 
In Fig.~\ref{fig:generalization_CT}, ``nnU-Net 0 shot" and MA-SAM 0 shot" denote that the models trained on BTCV data are directly used to perform inference on AMOS22 images, and ``nnU-Net 5 shot" and MA-SAM 5 shot" denote the models are further fine-tuned with 5 additional training cases from the AMOS22 dataset. 
From the results, we can see that our method exhibits better zero-shot and few-shot segmentation performance on AMOS22 CT and MRI images, demonstrating higher generalization capability.
Especially for MRI images, nnU-Net encounters complete failure in the zero-shot context, obtaining only 10.9\% Dice score, while our model still retains the performance of 60.4\% Dice score. 
In the five-shot context, our method also shows 9\% improvements than nnU-Net, further underscoring the advantages of generalization.

We also compare the model generalization on prostate MRI segmentation. 
In Table \ref{tab:generalization_prostate}, the results of nnU-Net and our models are obtained by directly applying the models fine-tuned on Site A to make predictions for Site B to F. 
We include two recent SOTA test-time domain generalization methods into comparison, i.e., TTST \citep{karani2021test} and TASD \citep{liu2022single}, which employ additional domain generalization techniques when performing predictions on each specific site. 
The results demonstrate that our method not only outperforms nnU-Net by a large margin when generalizing to different sites, but also achieves superior performance than SOTA domain generalization approaches. 
All these results on AMOS22 dataset and different prostate MRI datasets underscore the impressive generalization capability of our method, which is an importance characteristic for critical medical applications. 

\renewcommand{\arraystretch}{1.3}
\begin{table}[!t]
\centering
\caption{{Comparison of generalization performance of nnU-Net and our MA-SAM model with SOTA domain generalization methods on prostate datasets.}}
\resizebox{0.48\textwidth}{!}{%
\setlength\tabcolsep{2.0pt}
\scalebox{0.93}{
    \begin{tabular}{l|ccccc|c}
    \hline
          
        Methods &Site B &Site C &Site D &Site E &Site F &Average\\
        \hline   
        \hline
        
        nnU-Net \citep{isensee2021nnu}   &72.0 &69.6 &84.7 &42.5 &82.9 &70.3 \\
        TTST* \citep{karani2021test}    &86.0 &74.8 &81.0 &74.0 &80.9 &79.3 \\
        TASD* \citep{liu2022single}    &\textbf{87.1} &\textbf{76.4} &82.5 &76.0 &83.2 &81.1 \\
        \hline
        MA-SAM (Ours)    &{86.7} &{66.6} &\textbf{88.6} &\textbf{79.1} &\textbf{89.5} &\textbf{82.1} \\
        
        \hline
        
    \end{tabular}}}
\begin{tablenotes}
\item[*] \footnotesize{Note: $\ast$ means the method uses domain generalization techniques.}
\end{tablenotes}
\label{tab:generalization_prostate}
\end{table}

\begin{figure*}[!t]
\centering
\begin{minipage}[b]{.35\linewidth}
\centering
\captionsetup{width=.9\linewidth}
\captionof{table}{Comparison of model performance with different mask decoder designs.}
\vspace{-2mm}
\renewcommand{\arraystretch}{1.0}
    \begin{tabular}{l|c}
    \hline
          
        Decoder design &Dice [\%] \\
        \hline   
        \hline
        
        SAM mask decoder  &84.4 \\
        Progressive up-sampling &85.1 \\
        Multi-scale fusion &84.5 \\
        
    \hline
    \end{tabular}
    
    \label{tab:ablation_decoder}
\end{minipage}
\begin{minipage}[b]{.28\linewidth}
\centering
\captionsetup{width=.9\linewidth}
\captionof{table}{Comparison of model performance with different network backbones.}
\vspace{-2mm}
\renewcommand{\arraystretch}{1.0}
    \begin{tabular}{l|c}
    \hline
          
        ~~Backbone~~~~~~ &Dice [\%]\\
        \hline   
        \hline
        
        ~~ViT\_B  &82.5 \\
        ~~ViT\_L  &84.1 \\
        ~~ViT\_H  &85.1 \\
        
    \hline
    \end{tabular}
    \label{tab:ablation_backbone}
\end{minipage}
\begin{minipage}[b]{.35\linewidth}
\centering
\captionsetup{width=.9\linewidth}
\captionof{table}{Comparison of model performance with different position of 3D adapters.}
\vspace{-2mm}
\renewcommand{\arraystretch}{1.0}
    \begin{tabular}{l|c}
    \hline
          
        Position &Dice [\%] \\
        \hline   
        \hline
        
        Before MHSA  &86.7 \\
        After MHSA &86.8 \\
        Before \& after MHSA~~~~ &87.2 \\
        
    \hline
    \end{tabular}
    
    \label{tab:ablation_location}
\end{minipage}
\end{figure*}
\subsection{Ablation analysis of our method}
We conduct extensive ablation experiments on the BTCV dataset to investigate several key aspects regarding our SAM fine-tuning strategy: 1) effectiveness of each important component in our method, 2) effect of mask decoder design, 3) influence of network backbone, 4) choice of location for 3D adapters, 5) choice of rank for parameter-efficient fine-tuning. 

\renewcommand{\arraystretch}{1.1}
\begin{table}[]
\centering
\captionsetup{width=.9\linewidth}
\caption{Ablation on each key component in our method. The markers $\bullet$ and $\circ$ denote whether a specific component is used or not.}
\vspace{-2mm}
\resizebox{0.45\textwidth}{!}{%
\setlength\tabcolsep{2.0pt}
\scalebox{0.93}{
    \begin{tabular}{cccc|c}
    \hline

    SAM weights &~~~Full FT~~~    &~~~FacT~~~       &3D Adapters &Dice [\%] $\uparrow$ \\
    \hline
    \hline
    \Large$\circ$  &\Large$\bullet$ &\Large$\circ$   &\Large$\circ$      &72.2 \\
    \Large$\bullet$  &\Large$\circ$ &\Large$\circ$   &\Large$\circ$      &70.4 \\
    \Large$\bullet$&\Large$\bullet$ &\Large$\circ$   &\Large$\circ$      &85.3 \\
    \Large$\bullet$  &\Large$\circ$    &\Large$\bullet$ &\Large$\circ$   &85.1 \\
    \Large$\bullet$  &\Large$\circ$    &\Large$\circ$ &\Large$\bullet$   &86.4 \\
    \Large$\bullet$  &\Large$\circ$    &\Large$\bullet$ &\Large$\bullet$   &87.2 \\

    \hline
    \end{tabular}}}
    \label{tab:ablation_key_component}
\end{table}
\textit{1) Effectiveness of each component:}
We first validate the contribution of key components within our method, i.e., SAM's pre-trained weights, the parameter-efficient fine-tuning strategy with FacT, and the incorporation of 3D information with 3D adapters.
In Table~\ref{tab:ablation_key_component}, the ``Full FT" model denotes whether full fine-tuning is used for the image encoder, since the mask decoder is fully fine-tuned for all the models.
By comparing the results between the first and third rows in Table~\ref{tab:ablation_key_component}, we observe a substantial 13.1\% improvement in Dice when the model is initialized with SAM's pre-trained weights. This underscores the benefits of utilizing SAM's original weights that were pre-trained on a large-scale and diverse dataset. 
The second row shows that if the entire image encoder remains frozen without being fine-tuned, and only the mask decoder is updated, the performance is unsatisfactory. This suggests that due to the significant difference in image texture between natural and medical images, the image encoder trained solely on natural images struggles to extract essential features from medical images.
Moreover, Table~\ref{tab:ablation_key_component} demonstrates that FacT is capable of delivering performance on par with full fine-tuning, by adjusting a small portion of weight increments. 
The models equipped with 3D adapters achieve superior performance, validating the importance of incorporating the third-dimensional information for medical image segmentation.

\textit{2) Effect of mask decoder design:}
We compare the performance of different mask decoder designs, including the original SAM mask decoder, the progressive up-sampling strategy, and the multi-scale fusion strategy. 
Table \ref{tab:ablation_decoder} shows that the straightforward progressive up-sampling strategy yields superior results, validating its simplicity and effectiveness.
These results demonstrate the importance of recovering prediction resolution for medical images, which often contain small objects.
However, no significant improvements were observed with the multi-scale fusion strategy.
This might because of the extensive modifications it introduces to the original SAM decoder, resulting in less effective utilization of the pre-trained SAM weights.

\textit{3) Influence of network backbone:}
We conduct experiments with different network backbones, i.e., ViT\_B, ViT\_L, and ViT\_H, to assess their impact on the performance of our method.
As can be observed in Table \ref{tab:ablation_backbone}, there is a noticeable improvement in Dice performance as the model size increases from ViT\_B to ViT\_H, signifying the advantages of using a larger model size to enhance overall model performance.

\textit{4) Choice of location for 3D adapters:} 
We perform ablation experiments to investigate the placement of 3D adapters within our model. Specifically, we compare the performance when incorporating a 3D adapter in one of three locations: before the multi-head self-attention block (MHSA), after MHSA, or in both of these positions.
As demonstrated in Table~\ref{tab:ablation_location}, the configuration with two 3D adapters positioned both before and after MHSA yields superior performance for our final model.

\textit{5) Choice of rank:}
We investigate how the model's performance changes with varying decomposition rank $r$, by considering the rank value from the set \{4, 8, 16, 32, 64\}.
As expected, Table \ref{tab:ablation_rank} shows that with an increase in rank, there is a corresponding improvement in average Dice performance, but the performance tends to saturate when $r\geq 32$. 
We thus set $r=32$ in our experiments to seek a balance between performance gains and the number of parameters introduced.

\renewcommand{\arraystretch}{1.3}
\begin{table}[]
\centering
\caption{The change of Dice score for our method with different ranks.}
\vspace{-2mm}
\resizebox{0.45\textwidth}{!}{%
\setlength\tabcolsep{6.0pt}
\scalebox{0.93}{
    \begin{tabular}{l|ccccc}
    \hline
          
         &$r=4$ &$r=8$ &$r=16$ &$r=32$ &$r=64$\\
        \hline   
        \hline
        
        MA-SAM &81.4 &82.7 &84.6 &85.1 &85.3\\
        
        \hline
    \end{tabular}}}
    \label{tab:ablation_rank}
\end{table}
\section{Discussion}
Foundation models, like the Segment Anything Model (SAM), have revolutionized intelligent model development by offering robust generalization and few-shot learning capabilities. 
SAM has demonstrated impressive zero-shot performance for natural image tasks. However, applying SAM directly to medical image segmentation has proven ineffective due to the substantial domain differences. 
To address this problem, in this work, we propose a parameter-efficient fine-tuning method to adapt SAM to various medical imaging modalities.
Our method leverages FacT to efficiently update a small portion of weight increments and injects a set of designed 3D adapters to extract crucial volumetric or temporal knowledge of medical images during fine-tuning.
The general applicability and effectiveness of our method has been validated on four medical image segmentation tasks across three imaging modalities. 
Our model also demonstrates outstanding generalization capability, as well as significant advantage in particularly challenging tumor segmentation when prompts are used. 

One significant motivation for adapting SAM to medical images is its pre-training on a vast and diverse dataset, which is difficult to achieve in the filed of medical imaging. 
This makes SAM's adaptation generally applicable to various medical imaging modalities. 
In medical applications, there are recent efforts trying to pre-train modality-specific foundation models.
However, these models are often constrained to a specific medical imaging modality and challenging to extend to others.
For example, models pre-trained with chest x-ray data may face difficulties when applied to MRI data.
By leveraging SAM's pre-trained weights, we are able to train a large-scale segmentation network, such as ViT\_H, for medical image segmentation, even when limited data, such as just 5 imaging scans are used. 
Our experiments have demonstrated the benefits of increasing the model size, raising the intriguing question of how performance evolves with further increases in model size. Can we achieve improved accuracy or enhanced generalization with larger models for medical images? Exploring these possibilities holds great interest.

Using promptable segmentation is less meaningful for tasks that can already achieve satisfactory results with SOTA medical image segmentation methods.
Prompts prove particularly beneficial and valuable when dealing with challenging tumor segmentation tasks, as demonstrated in our experiments as well as other SAM fine-tuning works. 
However, crafting effective prompts demands a substantial amount of effort. As shown in Table~\ref{tab:comparison_tumor}, the performance of promptable segmentation drops as the quality of prompts declines. 
Given the challenges associated with manual prompt creation, there is considerate room for future exploration in automating this process. 
It would be interesting and valuable to investigate methods for generating suitable prompts automatically or study how to train an accurate segmentation model with noisy or imperfect prompts. 
This would enhance the practicality of promptable segmentation in scenarios where manual prompt creation is challenging.

\section{Conclusion}
We present an effective SAM adaptation framework that is general and can be applied to diverse medical image segmentation tasks across different modalities. 
Our method roots in the parameter-efficient fine-tuning strategy and successfully incorporates the volumetric or temporal information of medical images during fine-tuning.
Without using any prompt, our method with automatic segmentation outperforms various SOTA 3D medical image segmentation methods by a large margin. 
Our model has demonstrated outstanding generalization capability, which is crucial for successful deployment of intelligent model across medical datasets.
We have also shown the substantial advantage of the prompt mode, which is particularly valuable in tackling challenging tumor segmentation task.
Our method holds significant promise as a general segmentation framework that can be applied to various medical imaging modalities for both fully automatic and promptable segmentation.

\bibliographystyle{model2-names.bst}\biboptions{authoryear}
\bibliography{refs}

\end{document}